\documentclass[letterpaper, 10 pt, conference]{ieeeconf}  

\IEEEoverridecommandlockouts                              

\usepackage{graphicx} 
\usepackage{tabularx}
\usepackage{epsfig} 
\usepackage{mathptmx} 
\usepackage{times} 
\usepackage{amsmath} 
\usepackage{amssymb}  
\usepackage{multirow}

\def\IMGVSPACE{-0.3cm}

\title{\LARGE \bf 3DGS-ReLoc: 3D Gaussian Splatting for Map Representation and Visual ReLocalization
}

\author{Peng Jiang$^{1}$, Gaurav Pandey
$^{2}$ and Srikanth Saripalli$^{1}$
\thanks{$^{1}$ Peng Jiang and Srikanth Saripalli are with J. Mike Walker  \'{}66 Department of Mechanical Engineering, Texas A$\&$M University, College Station, TX-77843,USA {\tt\small maskjp,ssaripalli@tamu.edu}}%
\thanks{$^{2}$Gaurav Pandey is with The Department of Engineering Technology and Industrial Distribution Texas A$\&$M University, College Station, TX-77843, USA {\tt\small gpandey@tamu.edu}}%
}

\begin{document}

\maketitle
\thispagestyle{empty}
\pagestyle{empty}

\begin{abstract}
This paper presents a novel system designed for 3D mapping and visual relocalization using 3D Gaussian Splatting. Our proposed method uses LiDAR and camera data to create accurate and visually plausible representations of the environment. By leveraging LiDAR data to initiate the training of the 3D Gaussian Splatting map, our system constructs maps that are both detailed and geometrically accurate. To mitigate excessive GPU memory usage and facilitate rapid spatial queries, we employ a combination of a 2D voxel map and a KD-tree. This preparation makes our method well-suited for visual localization tasks, enabling efficient identification of correspondences between the query image and the rendered image from the Gaussian Splatting map via normalized cross-correlation (NCC). Additionally, we refine the camera pose of the query image using feature-based matching and the Perspective-n-Point (PnP) technique. The effectiveness, adaptability, and precision of our system are demonstrated through extensive evaluation on the KITTI360 dataset.
\end{abstract}

\section{introduction}
The rapid evolution of autonomous driving and robotic navigation technologies has underscored the critical importance of advanced scene reconstruction methodologies. These technologies rely heavily on the integration of data from diverse sensor modalities to create describable and accurate representations of the environment. Among various sensor fusion techniques, the combination of LiDAR and camera data is particularly noteworthy. This fusion harnesses LiDAR's precise depth sensing capabilities alongside the rich visual details captured by cameras, a synergy crucial for achieving the level of environmental understanding necessary for autonomous systems to navigate safely and efficiently. However, the challenge lies in harmonizing these different types of data into a unified, detailed, and geometrically accurate representation of the scene, a task that is both complex and essential for traversing intricate urban landscapes.

This paper introduces 3DGS-ReLoc, a novel system tailored for visual relocalization in autonomous navigation, employing 3D Gaussian Splatting (3DGS) as its primary map representation technique\cite{kerbl3DGaussianSplatting2023}. Utilizing LiDAR data, our method initiates the training of the 3D Gaussian Splatting representation, enabling the generation of large-scale, geometry-accurate maps. This initial training with LiDAR significantly improves our system's ability to create detailed and precise environmental models, which is essential for advanced perception systems in autonomous vehicles. Moreover, to address the high GPU memory consumption challenge, we adopt a strategy of dividing 3D Gaussian Splatting maps into 2D voxels and utilizing a KD-tree for efficient spatial querying.

3D Gaussian Splatting representation can generate high-fidelity images and depth data in association with known camera poses within the map's coordinates. This capability simplifies our method by facilitating the straightforward identification of correspondences between the query image and the Gaussian Splatting map through image feature detection and matching techniques. Additionally, by leveraging the depth information and its corresponding camera pose, we can accurately determine the camera pose of the query image. Implementing 3D Gaussian Splatting for visual relocalization not only showcases the adaptability of our method but also effectively tackles the complexities involved in fusing sensor data, contributing to the development of more precise and efficient scene representation techniques.

We conducted an extensive evaluation of our methodology with the KITTI360 dataset \cite{liaoKITTI360NovelDataset2023}. This dataset was chosen for its comprehensive annotations, which aid in creating accurate maps in diverse urban landscapes. Our results highlight our system's effectiveness, versatility, and precision. Specifically, we showcase the utility of 3D Gaussian Splatting for scene representation in visual relocalization tasks.

\section{related work}
\subsection{Map Representation}
Maps are crucial for robot navigation and autonomous driving, with traditional representations including voxel grids, point clouds, and meshes, as highlighted in recent literature \cite{chenDeepLearningVisual2023}. The advent of neural rendering techniques has introduced a new avenue for constructing maps with high fidelity. These models capture and depict 3D scenes by utilizing images and corresponding poses for guidance. This approach enables synthesizing high-fidelity images from novel views of the scene. Among these, Neural Radiance Fields (NeRF)\cite{mildenhallNeRFRepresentingScenes2020} has gained prominence. It encodes the radiance fields of complex 3-D scenes into the weights of multilayer perceptrons (MLPs), demonstrating exceptional realism in rendering 3-D environments through volume rendering under 2-D supervision. This innovation has significantly contributed to the development of mapping systems and the enhancement of SLAM (Simultaneous Localization and Mapping) systems, including iMAP\cite{sucarIMAPImplicitMapping2021} and NICE-SLAM\cite{zhuNICESLAMNeuralImplicit2022}. iMAP, for instance, employs an MLP for real-time scene representation within a SLAM framework, while NICE-SLAM introduces a dense, efficient, and robust SLAM approach by integrating multilevel local scene information and optimizing with geometric priors for better detail in large indoor scenes.

However, the scene-specific nature of networks trained with NeRF, where each 3-D scene's representation is encoded in an MLP's weights, restricts their generalizability across different environments. Furthermore, the computational intensity of NeRF-based methods results in slow rendering times. 3D Gaussian Splatting 
\cite{kerbl3DGaussianSplatting2023} has emerged as a viable alternative, providing an explicit representation more in line with traditional mapping approaches and enabling easier integration of conventional methods with minimal adjustments. This approach not only accelerates training times but also maintains high-quality visuals akin to NeRF. Recent efforts to apply 3D Gaussian Splatting to SLAM \cite{yugayGaussianSLAMPhotorealisticDense2023,matsukiGaussianSplattingSLAM2023,liSGSSLAMSemanticGaussian2024,keethaSplaTAMSplatTrack2023} have shown promise. SplaTAM \cite{keethaSplaTAMSplatTrack2023} represents an innovative application of 3D Gaussian splatting in SLAM, offering dense SLAM capabilities with monocular RGB-D cameras and enabling online camera pose tracking through singular 3D Gaussian Splatting map. Building on these efforts, \cite{liSGSSLAMSemanticGaussian2024} introduced SGS-SLAM, which incorporates 3D semantic segmentation into the GS-SLAM system. This method uses multi-channel optimization during mapping to combine appearance, geometric, and semantic constraints with key-frame optimization, enhancing the quality of reconstruction. Despite these advancements, the focus of research remains predominantly on indoor scenes of limited size, utilizing RGB-D cameras to generate dense point clouds. Several studies have been conducted on outdoor large-scale 3D Gaussian Splatting reconstruction \cite{linVastGaussianVast3D2024, chenPeriodicVibrationGaussian2023, yanStreetGaussiansModeling2024a, zhouDrivingGaussianCompositeGaussian2024}. However, these studies primarily focus on generating high-quality images \cite{linVastGaussianVast3D2024} or handling dynamic scenarios in street data \cite{zhouDrivingGaussianCompositeGaussian2024, yanStreetGaussiansModeling2024a} for simulation purposes, and do not explore their potential for map representation and relocalization.

\subsection{Visual Relocalization}
Visual relocalization aims to estimate a camera’s position and orientation from a single query image. Approaches to visual relocalization vary, including feature-based methods, scene coordinate regression, pose regression, and direct image alignment. DSAC \cite{brachmannDSACDifferentiableRANSAC2017} exemplifies the scene coordinate regression method, circumventing the need for explicit 3D mapping by mastering a pixel-to-point transformation through differentiable RANSAC for seamless end-to-end learning. In the realm of pose regression methods, the notable work by Laskar et al. \cite{laskarCameraRelocalizationComputing2017} stands out. They leverage Convolutional Neural Networks (CNNs) to identify similar images within a database and calculate relative poses, employing RANSAC to enhance accuracy. Meanwhile, PixLoc \cite{sarlinBackFeatureLearning2021} serves as a prime example of direct image alignment, utilizing deep multiscale features. PixLoc redefines localization as a metric learning challenge, facilitating comprehensive end-to-end training.

Despite the variety of methods, 2D-3D feature-based approaches remain predominant. 2D-3D feature-based approaches aims to estimate a camera's position and orientation (pose) from a 2D image within a previously mapped 3D scene. The construction of these 3D models typically involves Structure-from-Motion (SfM) with color images \cite{saputraVisualSLAMStructure2018}, Truncated Signed-Distance Function (TSDF) from range images \cite{wernerTruncatedSignedDistance2014}, or LiDAR-based mapping techniques \cite{wolcottVisualLocalizationLIDAR2014}. They compute the camera pose by matching 2D-3D correspondences through local feature descriptors. Since these descriptors often depend on the original imaging angle, research has focused on creating viewpoint-independent features \cite{detoneSuperPointSelfSupervisedInterest2018} or learning across different modalities, such as with P2-Net's unified descriptor for pixel-point matching \cite{wangP2NetJointDescription2021}. Contrastive learning has been explored to bridge the gap between camera images and LiDAR point clouds \cite{jiangContrastiveLearningFeatures2022b}. Additionally, approaches like that of Wolcott et al.\cite{wolcottVisualLocalizationLIDAR2014} propose localizing a camera within a 3D LiDAR-generated prior ground map by maximizing normalized mutual information between real camera measurements and generated synthetic LiDAR intensity image. Compared to traditional map representations, the 3D Gaussian Splatting representation has a more direct linkage between images, as it enables the rendering of new images and depth maps from novel viewpoints. This capability facilitates mitigating the challenges associated with view dependence, enhancing our ability to manage perspective-related difficulties more effectively.
\begin{figure*}
    \centering
    \includegraphics{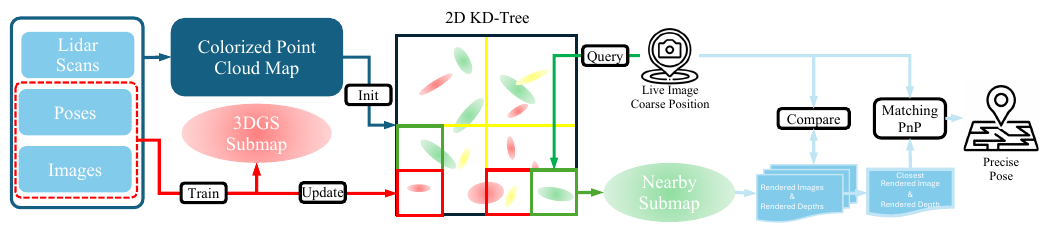}
    \caption{\textbf{Pipeline of 3D Gaussian Splatting for Map Representation and Visual ReLocalization}: The process starts by creating a colorized point cloud map from LiDAR scans, images, and poses. This map serves as the initialization for the 3D Gaussian Splatting (3DGS) map, which is incrementally trained on submaps. The 3DGS map is stored as a 2D voxel map, with a KD-tree enabling rapid spatial queries. For relocalization, a submap proximate to the query image's coarse pose is selected to render a series of images and depths. The query image is then subjected to a brute-force search against this image sequence to find the closest rendered image and depth. Subsequently, feature-based matching and the Perspective-n-Point (PnP) method are employed to iteratively refine the query image's pose, achieving precise localization within the global map.}
    \label{fig:pipeline}
\vspace{\IMGVSPACE}
\end{figure*}
\section{method}
This section will first revisit the concept of 3D Gaussian Splatting. Then, we will detail our system, which consists of two main components: the 3D Gaussian Splatting (3DGS) Map Representation and visual relocalization using the 3DGS Map. The complete system is illustrated in Fig.\ref{fig:pipeline}.
\subsection{Revisit 3D Gaussian Splatting}
The 3D Gaussian Splatting (3DGS) \cite{kerbl3DGaussianSplatting2023} is a rasterization technique designed for real-time rendering of photorealistic scenes using a group of 3D Gaussians for modeling. The original approach unfolds in three steps: a) employing the Structure from Motion (SfM) technique to estimate the poses of a collection of images from the same scene and a sparse point cloud of the scene; b) transforming each point in the cloud into a 3D Gaussian; c) applying Stochastic Gradient Descent (SGD) to refine the Gaussians, allowing for adaptive densification and pruning of the Gaussians based on the gradients and predefined criteria. The following parameters characterize each Gaussian in the model:
\begin{itemize}
    \item Center of the Gaussian  $\mu_i = \left[x_1,x_2,x_3\right] \in \mathbb{R}^3$, (usually initialize using sparse point cloud from SfM)
    \item Covariance matrix of the Gaussian $\Sigma_i=R_i S_i S_i^{\top} R_i^{\top}$\footnote{Covariance matrices are physically meaningful only when they are positive semi-definite. Constraining gradient descent to consistently produce such matrices is challenging, as update steps and gradients may unintentionally generate invalid matrices. As a resolution, the foundational paper\cite{kerbl3DGaussianSplatting2023} advocated for an alternate yet equally expressive optimization representation. Here, the covariance matrix $\Sigma_i$ analogously describes an ellipsoid's configuration, decomposed into scale and rotation matrices.}, comprised of a scaling matrix $S_i=\operatorname{diag}\left(\left[s_x, s_y, s_z\right]\right)$ and a rotation matrix $R_i=\mathrm{q}2\mathrm{R}\left(\left[r_w, r_x, r_y, r_z\right]\right)$, with $\mathrm{q}2\mathrm{R}$ converting a quaternion to a rotation matrix.
    \item RGB color  $c_i \in \mathbb{R}^3$ or spherical harmonics (SH) coefficients $c_i \in \mathbb{R}^k$, facilitating view-dependent colors with $k$ representing the degrees of freedom;
    \item Opacity $o_i\in \mathbb{R}$.
\end{itemize}
Accordingly, a 3D Gaussian  is defined as $g_i=\left[\mu_i,S_i,R_i,c_i,o_i \right]$ and a full 3DGS Map is a set of the Gaussian representation $G=\{g_0,...,g_N\}$

To render an image for a camera characterized by the intrinsic matrix  $K$ and pose $W_t$ (world-to-camera transformation), the Gaussians are first transformed into camera coordinates. They are then sorted by depth and rendered in a front-to-back sequence using Max's volume rendering formula \cite{maxOpticalModelsDirect1995a}:
\begin{equation}
C\left(\hat{x}\right)=\sum_{i \in \mathcal{S}} c_i q_i\left(\hat{x}\right) \prod_{j=1}^{i-1}\left(1-q_{j}\left(\hat{x}\right)\right)
\end{equation}
Here, the final rendered color $C\left(\hat{x}\right)$ at the camera projection plane for pixel $\hat{x}$ is the weighted sum of each Gaussian's color $c_i$. The weight is calculated using the footprint function $q_{i}$ derived from the Gaussian kernel \cite{zwickerEWAVolumeSplatting2001} (see Eq.\ref{eq:footprint}), and is modulated by an occlusion (transmittance) term that accounts for all Gaussians preceding the current one.
\begin{equation}
q_{i}(\hat{x}) =o_i\frac{1}{\left|J^{-1}\right|\left|W^{-1}\right|}G_{\hat{\Sigma}_i^c}\left(\hat{x}-\hat{\mu}_i\right)
\label{eq:footprint}
\end{equation}
where $G_{\hat{\Sigma}_i^c}$ is a Gaussian function with covariance matrix $\hat{\Sigma}_i^c$, a $2\times2$ matrix obtained by excluding the last row and column from the matrix computed using Eq.\ref{eq:foot_conv}, and $\hat{\mu}=[x_1,x_2]$ is the first two value of mean $\mu$ of this Gaussian.
\begin{equation}
   \Sigma_i^c=J W \Sigma_i W^{\top} J^{\top}
\label{eq:foot_conv} 
\end{equation}
Where $J=\partial m\left(\mu\right) / \partial \mu$ is the Jacobian of the projection formula Eq. \ref{eq:proj}:
\begin{equation}
m\left({\mu}\right)=K\left(\frac{W \mu}{(W \mu)_z}\right)
\label{eq:proj}
\end{equation}
For a comprehensive derivation of the footprint function, readers are directed to \cite{zwickerEWAVolumeSplatting2001}.

For rendering depth, we can simply replace the color $c_i$ with the $z_i=x_3$ of the Gauassian transformed in the camera coordinate:
\begin{equation}
D\left(\hat{x}\right)=\sum_{i \in \mathcal{S}} z_i q_i\left(\hat{x}\right) \prod_{j=1}^{i-1}\left(1-q_{j}\left(\hat{x}\right)\right)
\end{equation}
\subsection{3D Gaussian Splatting Map Representation}
\subsubsection{3D Map Construction and Initialization}
Contrary to the original methods that begin with Structure from Motion (SfM) as outlined in \cite{kerbl3DGaussianSplatting2023}, our approach initiates the 3D Gaussian process by utilizing the 3D map generated from LiDAR point cloud data and corresponding images. This ensures that our foundational representation possesses accurate geometric information.
However, the depth information derived from the LiDAR point cloud is sparse, presenting challenges for subsequent visual localization tasks due to the possibility that keypoints detected at this stage may lack corresponding LiDAR depth. Nevertheless, by leveraging the densification scheme of the 3D Gaussian Splatting method \cite{kerbl3DGaussianSplatting2023}, our method can increase the number of underlying Gaussians during the training process. This enhancement allows rendering dense depth from various viewpoints using 3DGS representation. This dense depth can provide precise depth information for our visual localization tasks.
\subsubsection{Map Storage and Management}
The 3D Gaussian Splatting method is known for its high GPU memory consumption, making the representation of large outdoor scenes challenging. To mitigate this, we have opted to use only RGB color information, foregoing the use of Spectral Harmonics (SH) decomposition. While Spherical Harmonics (SH) decomposition aids in capturing lighting and view-dependent effects, it significantly raises memory requirements. Our choice reduces the map's memory footprint, but it constrains direct comparisons of rendering quality with methods encoding lighting information—key in outdoor settings due to complex light and shadow dynamics (see Section \ref{sec:dis_vq}). Our primary aim is establishing a dependable mapping system for visual relocalization, making detailed rendering quality comparisons, particularly regarding lighting effects, beyond this work's scope.

To efficiently manage and train large-scale 3DGS maps, we've organized the 3D environment into a 2D voxel grid. This method divides the map into smaller voxels, each storing 3DGS parameters based on the $\mu$, and assigns a unique hash ID to every voxel for quick querying. For better efficiency in spatial querying and updating 3DGS parameters in voxels according to the camera pose of each image, we utilize a KD tree. Constructed from the voxels' center points, the KD tree swiftly identifies voxels within a certain range of the cameras. This approach not only improves our system's scalability but also minimizes the use of computational resources, allowing for the detailed reconstruction of large environments without excessive GPU memory demands.
\subsubsection{Loss Function}
The original 3D Gaussian Splatting technique was developed primarily for novel view synthesis, focusing on producing high-quality images. Consequently, it utilizes a balanced combination ($L_{rgb}$) of the Mean Absolute Error loss ($L_1$) and the Structural Similarity Index Measure (SSIM) loss ($L_{D-SSIM}$) to evaluate the difference between rendered ($\left(\mathcal{R}\left(G_t\right)\right)$) image and actual ground truth image ($I_t$) (see to Eq. \ref{eq:photo}). This singular focus allows Gaussian Splatting to compromise the precision of rendered depth in favor of enhancing the visual quality of RGB images. Additionally, this leads the densification process to introduce new Gaussians that do not adhere to the underlying geometry.
\begin{equation}
L_{rgb}(I_1,I_2)=(1-\lambda) L_1(I_1,I_2)+\lambda L_{\text{D-SSIM}}(I_1,I_2)
\end{equation}
\begin{equation}
L_{photo} =L_{rgb}\left(\mathcal{R}\left(G_t\right), I_t\right)
\label{eq:photo}
\end{equation}
To address this limitation, we incorporate a re-projection error loss aimed at maintaining both the geometric accuracy of the scene representation and the fidelity of rendered depth. We acquire depth information ($\mathcal{D}\left(G_t\right)$) at pose $W_t$ through Gaussian Splatting ($G_t$). This depth information is then used to re-project the ground truth image ($I_t$) from pose $W_t$ to pose $W_{t+1}$ through transformation matrix $T_{t}^{t+1}$, and the re-projected image ($\mathcal{P}\left(\mathcal{D}\left(G_t\right),I_{t},T_{t}^{t+1}\right)$) is subsequently compared with the actual ground truth image ($I_{t+1}$) at pose $W_{t+1}$ (see Eq. \ref{eq:reproj}).
\begin{equation}
L_{reproj} = L_{rgb}\left(\mathcal{P}\left(\mathcal{D}\left(G_t\right),I_{t},T_{t}^{t+1}\right), I_{t+1}\right)
\label{eq:reproj}
\end{equation}
 As a result, our comprehensive loss function is outlined in Eq. \ref{eq:total}, with $\omega_1$ and $\omega_2$ serving as weights to balance the contributions of the two loss functions effectively.
\begin{equation}
L = \omega_1L_{photo}+\omega_2L_{reproj}
\label{eq:total}
\end{equation}

\subsection{Visual ReLocalization Method Using 3DGS Map}
\subsubsection{Initial ReLocalization}\label{sec:method_init_loc}
Our approach starts with leveraging raw pose data to pinpoint the query's location on the global map. This data may come from various sources, including GPS systems. Utilizing the raw pose as a reference, we retrieve a segment of the global 3DGS map most likely to contain the query image's precise location.

After selecting the nearby submap, we refine location accuracy through a brute-force search. This involves generating and comparing multiple images from the 3DGS submap with the query image to find the most visually similar one, assuming that similarity indicates closeness in location. This method improves upon GPS accuracy, providing a precise starting point for feature-based matching. We employ normalized cross-correlation (NCC) \cite{hiasa2018cross} for this image comparison,  a metric frequently applied in medical image registration to evaluate similarity, as defined below:
\begin{equation}
\mathrm{NCC}(I_q, I_{G_{t}})=\frac{\sum_{(i, j)}(I_q-\bar{I}_q)(I_{G_{t}}-\bar{I}_{G{t}})}{\sqrt{\sum_{(i, j)}(I_q-\bar{I}_q)^2} \sqrt{\sum{(i, j)}(I_{G_{t}}-\bar{I}_{G{t}})^2}}
\label{eq:ncc}
\end{equation}

Where $I_q$ represents the query image, $I_{G_{t}}$ denotes the image rendered from the 3DGS submap at pose $W_t$, and $\bar{I}$ indicates the mean intensity of the images. 

Another rationale behind selecting normalized cross-correlation is its differentiable nature, which aligns well with the differentiable characteristics of the 3D Gaussian representation. This compatibility has the potential to facilitate a fully differentiable relocalization pipeline, creating avenues for seamless integration and optimization within the overall system. (For a more discussion, please see Section \ref{sec:dis_diff}).
\subsubsection{ReLocalization Refinement}\label{sec:feat_match}
After pinpointing the closest rendered image, we adopt a feature-based matching technique. This phase entails identifying matching points between the query image and the closest rendered counterpart. By harnessing the known camera pose associated with the rendered image, along with the depth rendered from the 3DGS map, we employ the Perspective-n-Point (PnP) algorithm to refine the pose of the query image within the selected submap. 

To further enhance the precision of localization, we employ an iterative refinement process on the initially estimated pose. This involves repeatedly performing the feature-based matching step with images newly rendered using the pose estimated from the preceding step. Each cycle is designed to progressively refine the pose estimation, capitalizing on the increased accuracy with each iteration to achieve the most precise localization achievable. 

Considering the broad spectrum of available feature detection and matching algorithms\cite{saputraVisualSLAMStructure2018}, we opted for Superpoint \cite{detoneSuperPointSelfSupervisedInterest2018} for keypoint detection and feature extraction, coupled with LightGlue \cite{lindenbergerLightGlueLocalFeature2023} for the matching process. These choices were driven by their proven effectiveness and compatibility with our localization framework, enabling us to achieve high-quality feature matching and efficient pose recovery, as shown in Section \ref{sec:exp}.
\begin{figure*}[t]
\centering
\vspace*{0.2cm}
\begin{tabular}{cc} 
\includegraphics[width=3.3in]{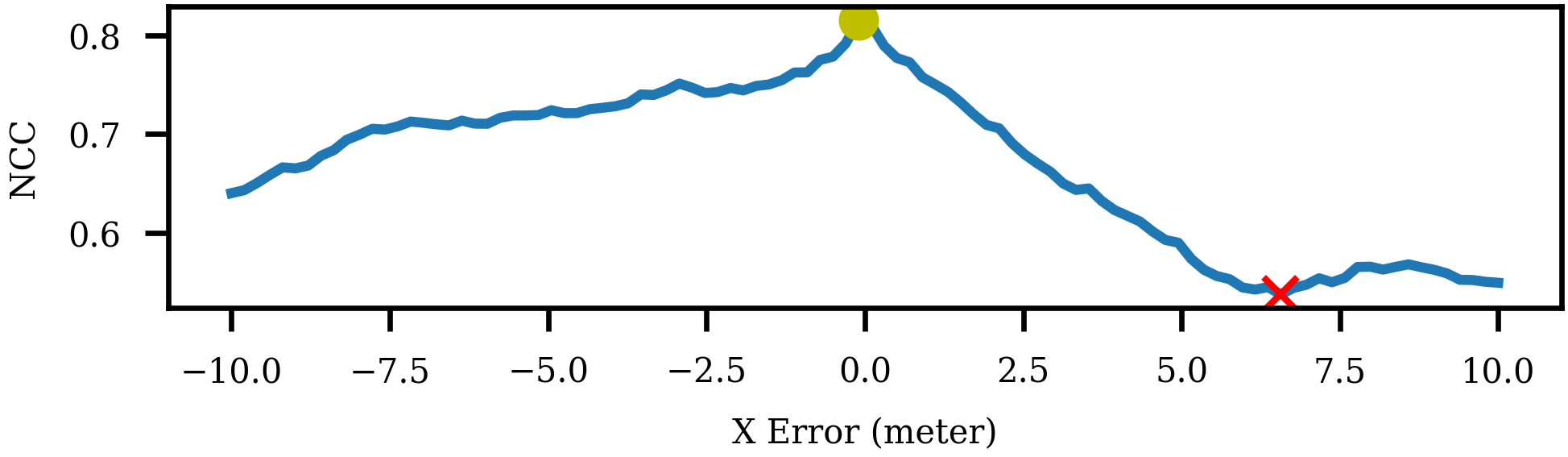}&\includegraphics[width=3.3in]{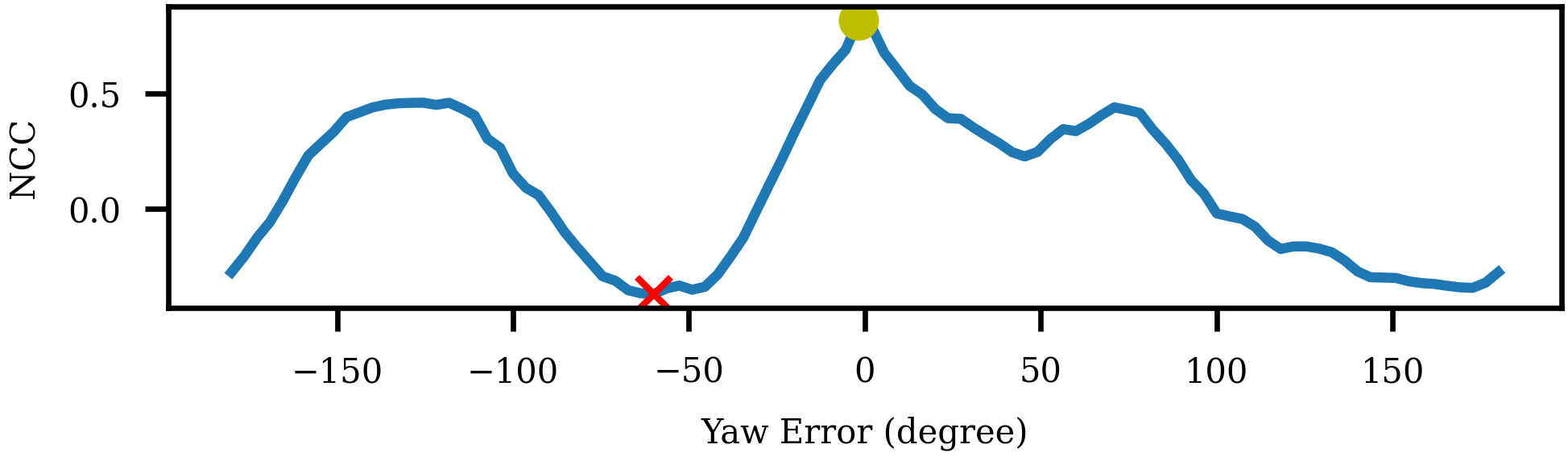}\\
{\scriptsize(a) X Error-Normalized Cross Correlation}&{\scriptsize(e) Yaw Error-Normalized Cross Correlation}\\
\includegraphics[width=3.3in]{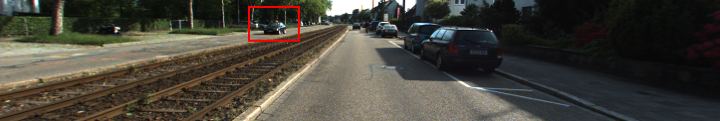}&\includegraphics[width=3.3in]{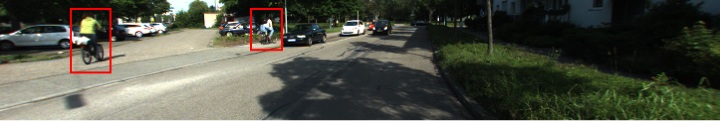}\\
{\scriptsize(b) Query Image}&{\scriptsize(f) Query Image}\\
\includegraphics[width=3.3in]{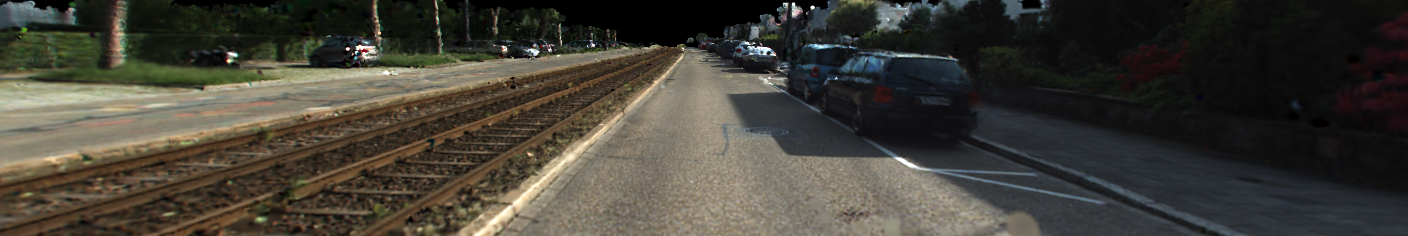}&\includegraphics[width=3.3in]{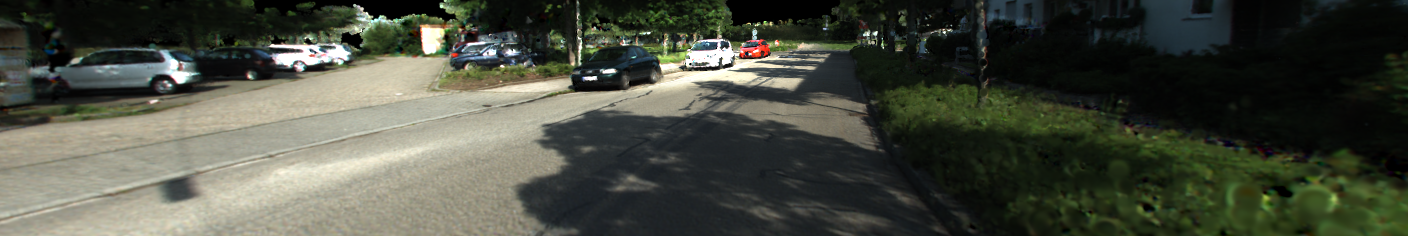}\\
{\scriptsize(c) Best Matched Rendered Image Along X}&{\scriptsize(g) Best Matched Rendered Image Along Yaw}\\
\includegraphics[width=3.3in]{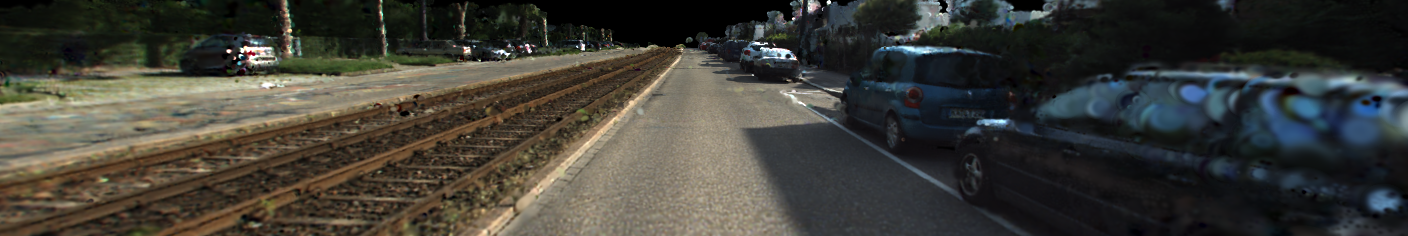}&\includegraphics[width=3.3in]{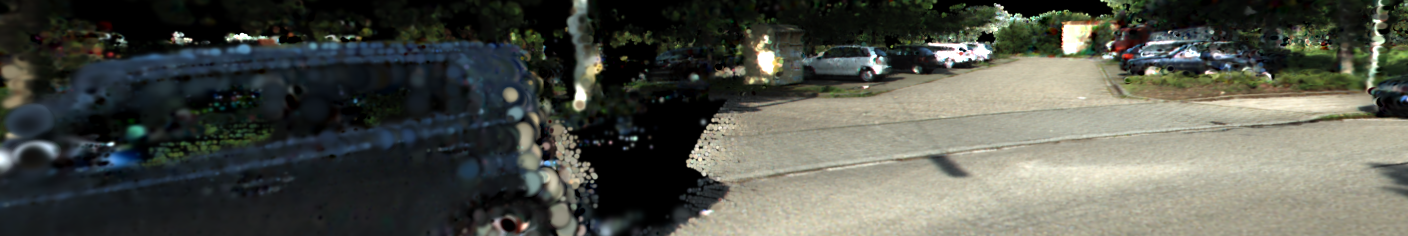}\\
{\scriptsize(d) Worst Matched Rendered Image Along X}&{\scriptsize(h) Worst Matched Rendered Image Along Yaw}\\
\end{tabular} 
\caption{(a)/(e) Illustrating the Relationship between X/Yaw Error and Normalized Cross-Correlation in Localization Initialization; (b)/(f) Query Image for Localization; (c)/(g) Best Matches in Rendered Image Sequences; (d)/(h) Worst Matches in Rendered Image Sequences.} 
\label{fig:init_loc_xy}
\vspace{\IMGVSPACE}
\end{figure*}

\subsubsection{Live Relocalization}\label{sec:method_live_reloc}
In live relocalization task, the system must continuously track a camera's pose using streaming images. For the initial query image, we conduct initialization and refinement as outlined in Sections \ref{sec:method_init_loc} and \ref{sec:feat_match}. For images that follow, we adopt a constant velocity model for predicting the camera's next pose, further refining the pose with the feature-matching technique described in Section \ref{sec:feat_match}. This streamlined approach eliminates the necessity of brute-force searches for every query image, significantly boosting the efficiency of ongoing relocalization efforts. Such efficiency is crucial for real-time applications in fields like autonomous navigation and robotics, providing smoother and more reliable tracking. To enhance pose estimation accuracy, we can incorporate more sophisticated methods, such as filter-based techniques \cite{thrunProbabilisticRoboticsIntelligent2005}.
\section{experiment}\label{sec:exp}
Our experimental evaluation was conducted using the KITTI360 dataset, which includes LiDAR data, camera images, ground truth poses, and semantic/instance labels. We focus on the dataset's first route (\textit{2013\_05\_28\_drive\_0000\_sync}), dividing it into two segments: the initial drive and the revisit one.

The 3D Gaussian Map was constructed using data solely from the first drive, which provided the necessary inputs for initializing and training our 3DGS map. This map aimed to establish a reliable reference for our relocalization tasks. Data from the second drive were then used to test our relocalization algorithm, allowing us to measure our approach's effectiveness in a real-world setting. Specifically, we selected two subsequences for our experiments: 
\begin{itemize}
    \item Seq0: frames 4200 to 4550 were used for map creation, and frames 7779 to 8002 for visual relocalization;
    \item Seq1: frames 7120 to 7450 were used for constructing the map, and frames 9754 to 10062 for visual relocalization;
\end{itemize}

During map construction, we utilized the dataset's semantic annotations to exclude the sky and instance labels corresponding to moving objects like vehicles and pedestrians. This approach concentrated our efforts on static environmental features, eliminating the necessity for dynamic reconstruction within the 3DGS model \cite{luitenDynamic3DGaussians2023}. While our mapping system is capable of processing the entire route, our relocalization experiments focused on selected subsequences. All training and experimental activities were conducted on an NVIDIA RTX A4000 equipped with 16 GB of memory. The submap size was set to 120 meters for training and 150 meters for relocalization, with a voxel resolution of 1 meter.

\subsection{Initial ReLocalization}\label{sec:exp_init_loc}
In the initial phase of relocalization, we employ the normalized cross-correlation (NCC) metric to evaluate the similarity between pairs of images as mentioned in Section \ref{sec:method_init_loc}. To illustrate the utility of NCC, we present two examples demonstrating its effectiveness in overcoming common localization challenges.

In Fig. \ref{fig:init_loc_xy} (a-d), we examine the scenario where there is an error in the yaw angle. Fig. \ref{fig:init_loc_xy}(b) displays the query image used for relocalization.  Despite the presence of two new bicyclists in the query image, the NCC metric successfully identifies the correct match (indicated by the yellow point), demonstrating robustness to changes in scene composition and minor orientation errors.

Further, Fig. \ref{fig:init_loc_xy} (e-h) explores the impact of errors in the x position on the localization process. Remarkably, the NCC metric maintains its effectiveness even with an error margin of up to 10 meters, accurately localizing the correct position. This scenario reveals a notably negative relationship between the x position error and the NCC metric’s performance, underscoring the metric's capacity to guide correct localization under significant positional discrepancies.

These examples highlight the NCC metric's effectiveness in handling orientation (yaw) and positional (x) errors during initial relocalization. Utilizing the NCC metric enhances our method's ability to withstand scene variations and inaccuracies in starting positions, laying a robust groundwork for accurate localization in complex settings. It's important to note that the closely matched examples presented here benefit from the use of a very fine grid size during the search process. However, in practical applications and in our implementation, we opt for a coarser grid size to expedite the initialization phase. For achieving precise relocalization, we subsequently apply a feature-based matching method, as detailed in Section \ref{sec:feat_match} and illustrated in the subsequent section.

\begin{table}[]
\centering
\vspace{0.2cm}
\resizebox{\columnwidth}{!}{%
\begin{tabular}{cccccc}
\hline
Seq             & Success                  & Stage  & X Error    &  Y Error   &  Yaw Error    \\ \hline
\multirow{2}{*}{0} & \multirow{2}{*}{219/223} & Init  & 3.513 (3.080) & 2.381 (1.807) & 14.007 (10.685)    \\ \cline{3-6} 
                   &                          & Refine & 0.185 (0.189) & 0.117 (0.168) & 0.535 (0.498) \\ \hline
\multirow{2}{*}{1} & \multirow{2}{*}{304/308} & Init   & 3.212 (2.535) & 3.148 (2.450) & 12.001 (13.388)    \\ \cline{3-6} 
                   &                          & Refine & 0.098 (0.076) & 0.114 (0.103) & 0.247 (0.239) \\ \hline
\end{tabular}
}
\caption{Evaluation for Single Image Query Re-localization Error in Initialization and Refinement Stage }
\label{tabel:eval_single}
\vspace{\IMGVSPACE}
\end{table}

\begin{figure}[t]
\vspace{0.2cm}
\centering
\begin{tabular}{cc} 
\includegraphics[width=1.6in]{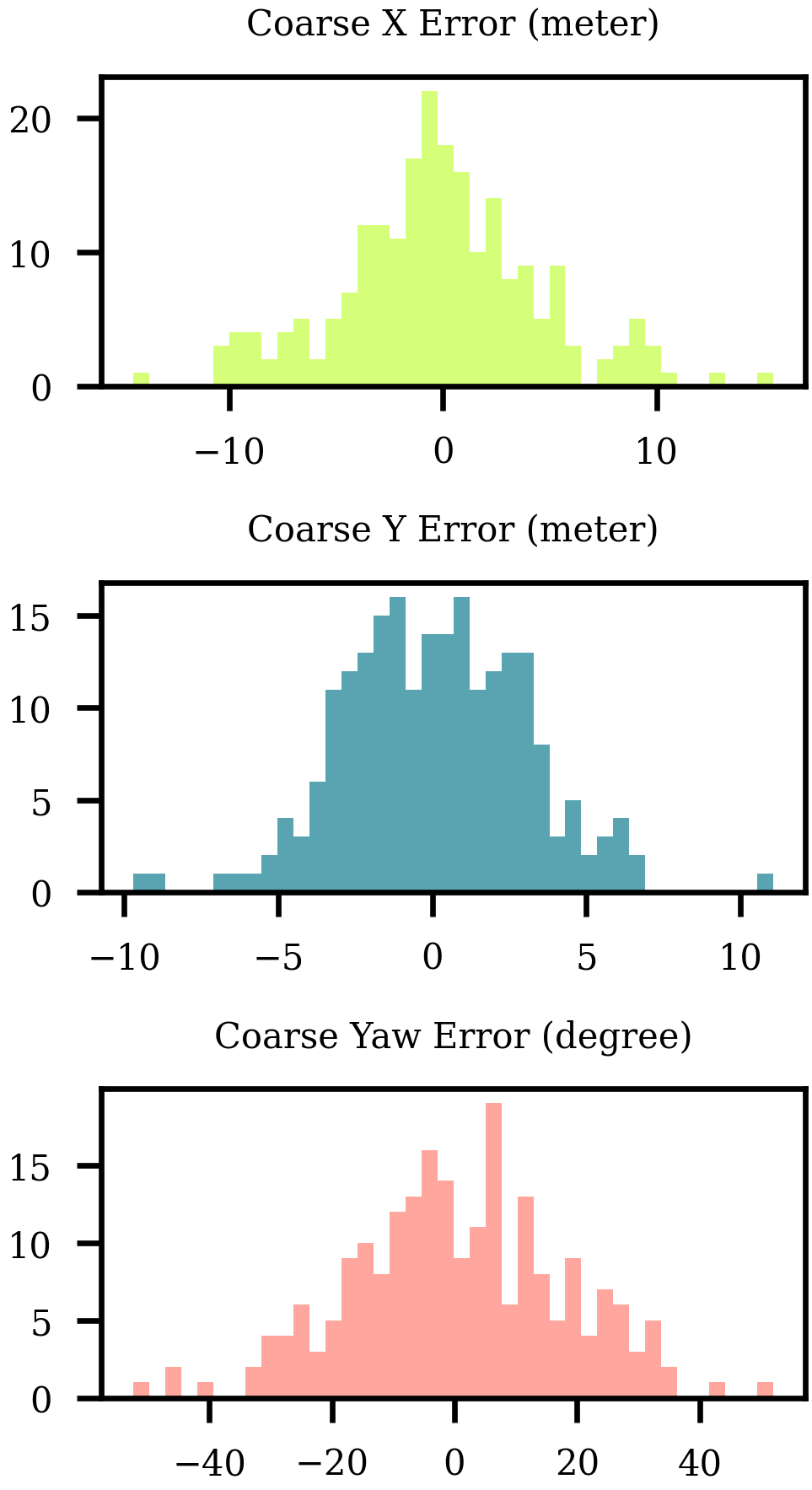}&\includegraphics[width=1.6in]{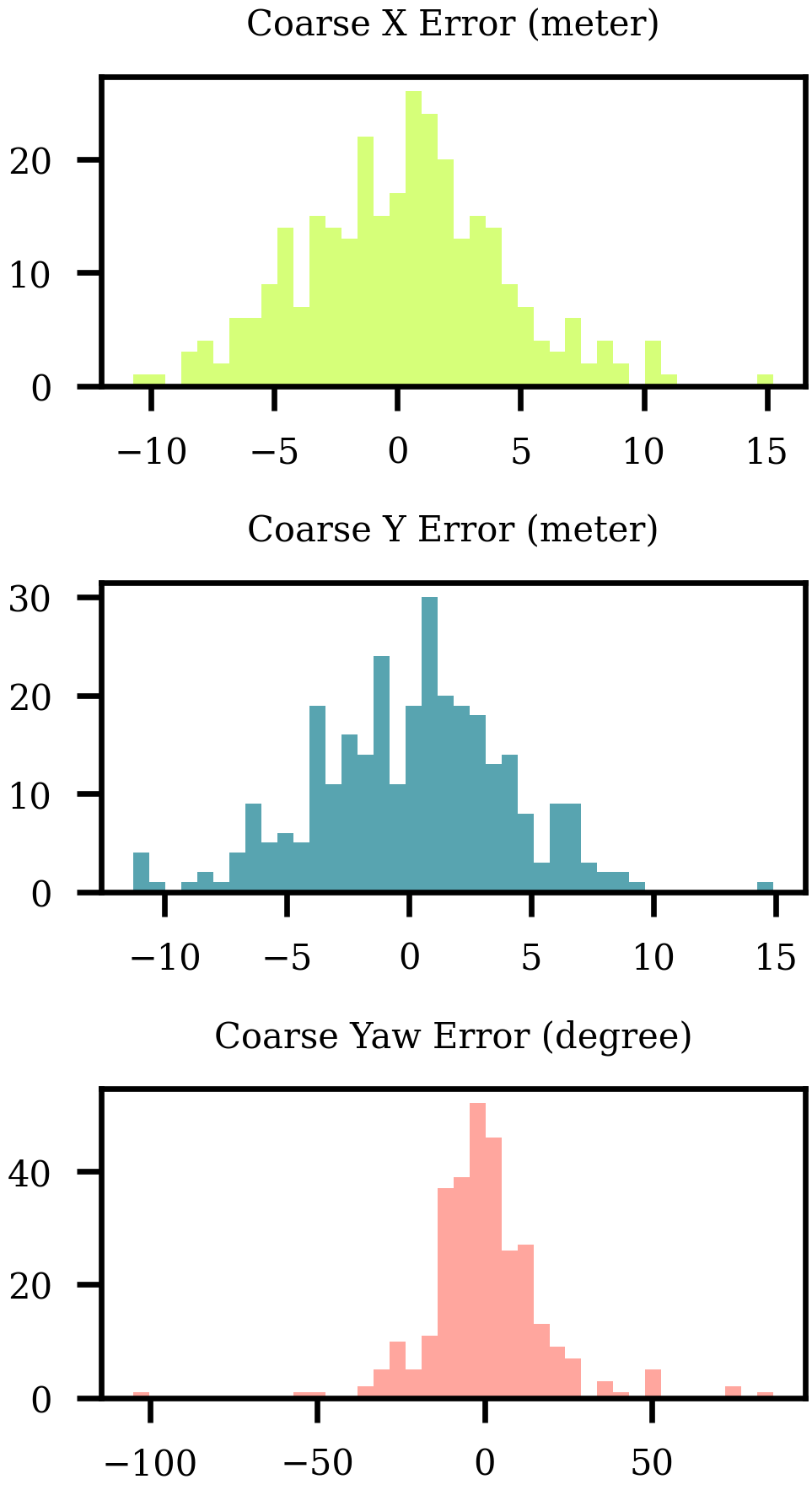}\\
{\scriptsize(a) Seq 0}&
{\scriptsize(b) Seq 1}\\
\end{tabular} 
\caption{Evaluation of Initial Localization X, Y, Yaw Error Histogram} 
\label{fig:init_loc_seq}
\vspace{\IMGVSPACE}
\end{figure}

To evaluate our method's effectiveness more thoroughly, we used all query images for the initial relocalization analysis. We introduced noise into $(x, y, yaw)$ of the ground truth pose of each query image. The noise was uniformly sampled within a range of $(-10,10)$ meters for the $x$ and $y$ translations and $(-90^{\circ},90^{\circ})$ for the $yaw$ rotation. A brute-force search was conducted with a grid size of $2$ meters and $10^{\circ}$ within a search space of $(-15,15)$ meters and $(-360^{\circ},360^{\circ})$.We explored the initial $20\%$ of the search space using a random search and applied an early stop criterion. This criterion was based on whether the Normalized Cross-Correlation (NCC) dropped below a set threshold and whether we could successfully obtain sufficient matching points with the second-stage method. The outcomes of this evaluation are detailed in Table \ref{tabel:eval_single} and illustrated in Figure \ref{fig:init_loc_seq}. As indicated in Table \ref{tabel:eval_single}, both sequences exhibit high success rates, with Seq 0 achieving a 98.2\% success rate (219 out of 223 attempts) and Seq 1 achieving a 98.7\% success rate (304 out of 308 attempts). Following the exclusion of unsuccessful matches, we calculated the mean and standard deviation of the errors in $(x,y,yaw)$. The distribution of translation errors, predominantly within the $(-5,5)$ meter range yet occasionally exceeding this, is depicted in Figure \ref{fig:init_loc_seq}. Despite the presence of significant translation errors initially, the refinement stage markedly enhanced localization accuracy. 

\subsection{ReLocalization Refinement}
\begin{figure}[t]
\vspace{0.2cm}
\centering
\begin{tabular}{cc} 
\includegraphics[width=1.6in]{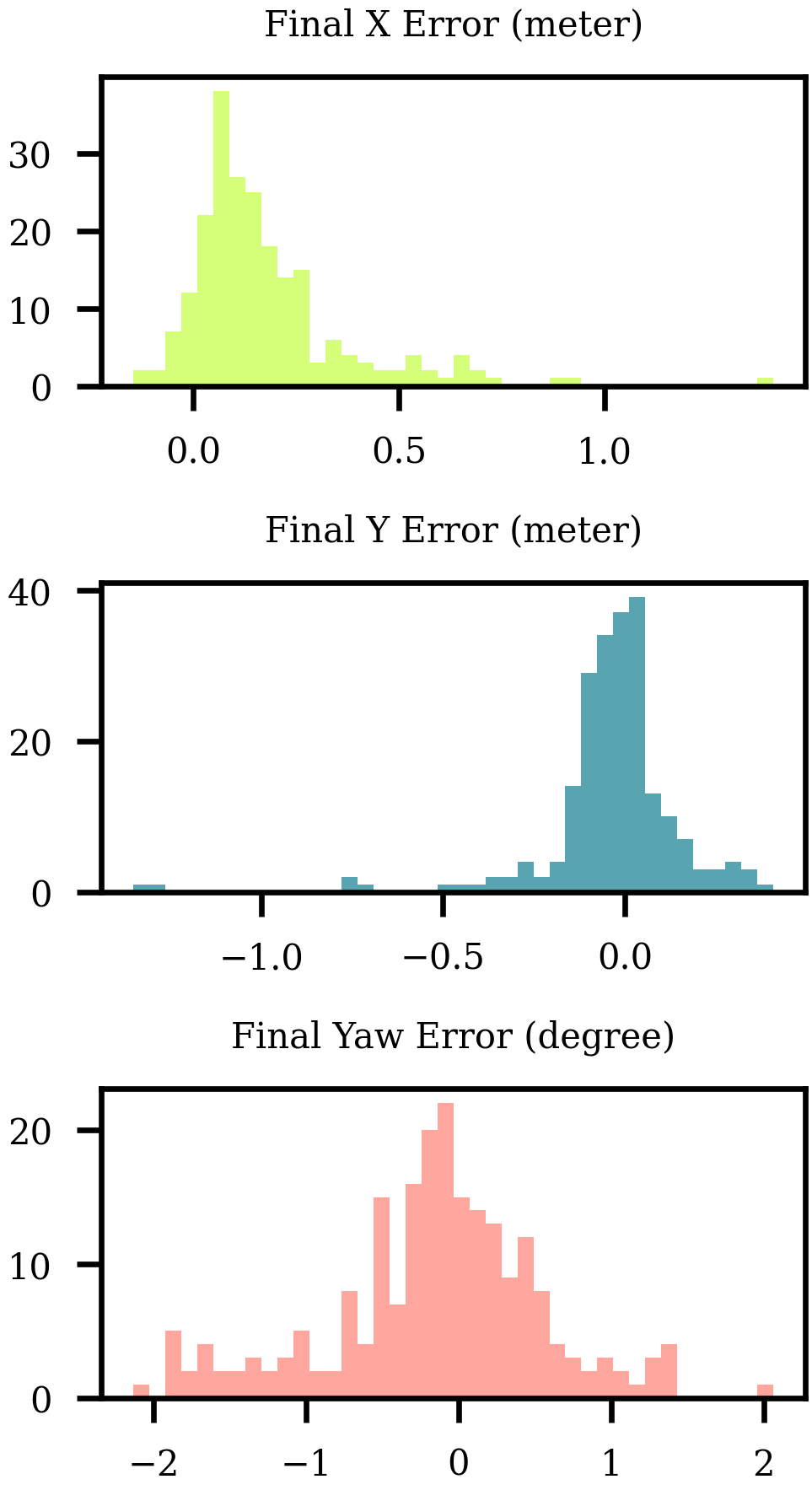}&\includegraphics[width=1.6in]{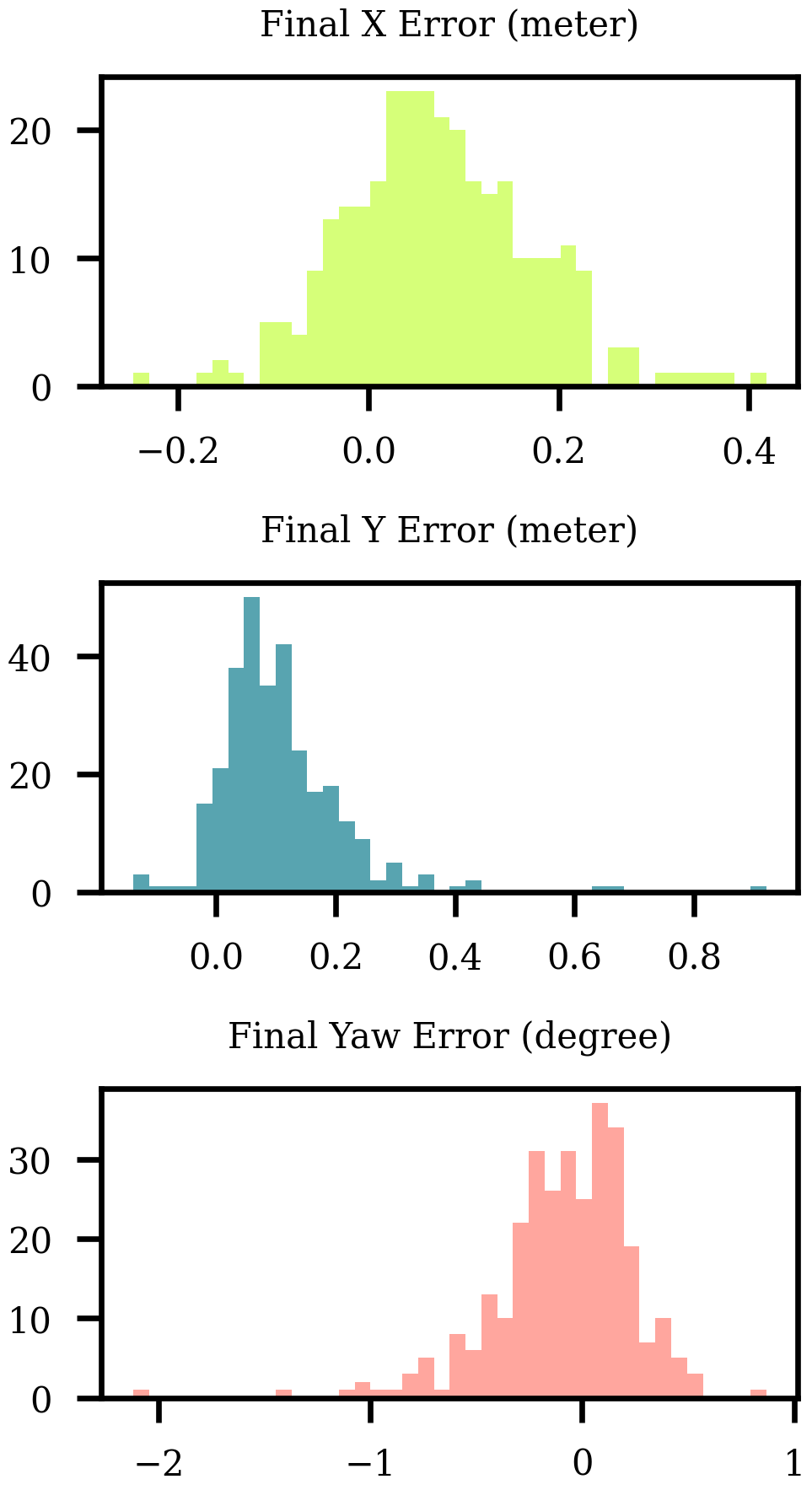}\\
{\scriptsize(a) Seq 0}&
{\scriptsize(b) Seq 1}\\
\end{tabular} 
\caption{Evaluation of Re-Localization X, Y, Yaw Error Histogram} 
\label{fig:refine_loc_seq}
\vspace{\IMGVSPACE}
\end{figure}

To thoroughly assess the accuracy of our final fine-pose estimations, we started with the initial poses that were successfully obtained, as outlined in Section \ref{sec:exp_init_loc}. These poses underwent processing via keypoint detection and feature extraction utilizing Superpoint \cite{detoneSuperPointSelfSupervisedInterest2018}, followed by matching through LightGlue \cite{lindenbergerLightGlueLocalFeature2023}. This cycle of detection, description, and matching was iteratively performed to enhance the accuracy of our estimations. The outcomes of these iterative enhancements are succinctly summarized in Table \ref{tabel:eval_single} and illustrated in Figure \ref{fig:refine_loc_seq}.

Following the refinement process, we observed significant improvements in the results. For instance, in Seq 0, initial positioning errors decreased from $3.513$ meters to $0.185$ meters in the X-axis, from $2.381$ meters to $0.117$ meters in the Y-axis, and from $14.007^{\circ}$ to $0.535^{\circ}$ in $Yaw$. Similarly, in Seq 1, errors were reduced from $3.212$ meters to $0.098$ meters in $X$, from $3.148$ meters to $0.114$ meters in $Y$, and from $12.001^{\circ}$ to $0.247^{\circ}$ in $Yaw$. Beyond the reduction in errors, the consistency of our methodology is also evident from the diminished standard deviation values, showcasing the reliability and precision of our approach.

\subsection{Live ReLocalization}
For live relocalization evaluation, we randomly initialized the pose of the first query image, which corresponds to the first frame in each sequence. We then streamed subsequent images for live relocalization.
As detailed in Table \ref{table:eval_traj}, we utilize Absolute Pose Error (APE) and Relative Pose Error (RPE) to evaluate our system's performance on live relocalization task. The table presents comprehensive statistics for both metrics across two sequences, including the Root Mean Square Error (RMSE), Mean, Median, Standard Deviation (Std), Minimum (Min), Maximum (Max), and Sum of Squared Errors (SSE).
\begin{table}[]
\centering
\vspace{0.2cm}
\resizebox{\columnwidth}{!}{%
\begin{tabular}{ccccccccc}
\hline
Metric              & Seq & RMSE  & Mean  & Median & Std   & Min   & Max   & SSE   \\ \hline
\multirow{2}{*}{APE} & 0   & 0.103 & 0.092 & 0.088  & 0.047 & 0.013 & 0.349 & 2.387 \\ \cline{2-9} 
                     & 1   & 0.099 & 0.087 & 0.079  & 0.047 & 0.013 & 0.311 & 3.032 \\ \hline
\multirow{2}{*}{RPE} & 0   & 0.083 & 0.070 & 0.060  & 0.046 & 0.008 & 0.252 & 1.543 \\ \cline{2-9} 
                     & 1   & 0.083 & 0.070 & 0.060  & 0.045 & 0.008 & 0.292 & 2.140 \\ \hline
\end{tabular}
}
\caption{Evaluation for Live Re-localization using Absolute Pose Error (APE) and Relative Pose Error (RPE)}
\label{table:eval_traj}
\vspace{\IMGVSPACE}
\end{table}

For APE, RMSE is around $0.1$, with an average error near $0.09$ and a median of $0.08$, indicating high accuracy with low variability (standard deviation of $0.047$). SSE values highlight precise pose estimations over time. 

RPE shows consistent metrics for sequences 0 and 1, with an RMSE of $0.083$ and mean and median errors of $0.070$ and $0.060$, respectively, showing stable relative pose accuracy. Standard deviations are minimal ($0.046$ for Sequence 0 and $0.045$ for Sequence 1), with error ranges of $0.008$ to $0.252$ for Sequence 0 and $0.008$ to $0.292$ for Sequence 1, and SSE values of $1.543$ for Sequence 0 and $2.140$ for Sequence 1, indicating robust relative pose estimation.

\begin{figure}[t]
\centering
\begin{tabular}{cc} 
\includegraphics[width=1.6in]{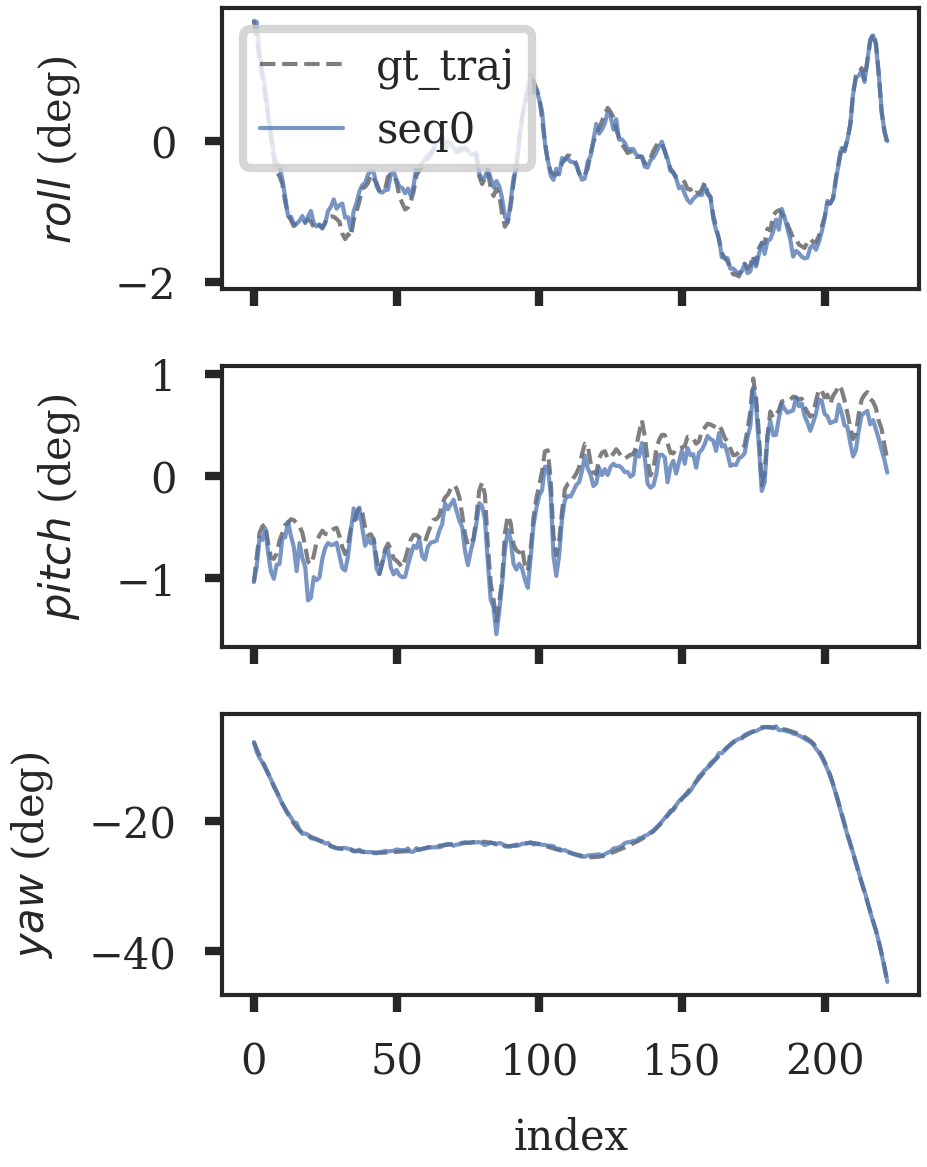}&\includegraphics[width=1.6in]{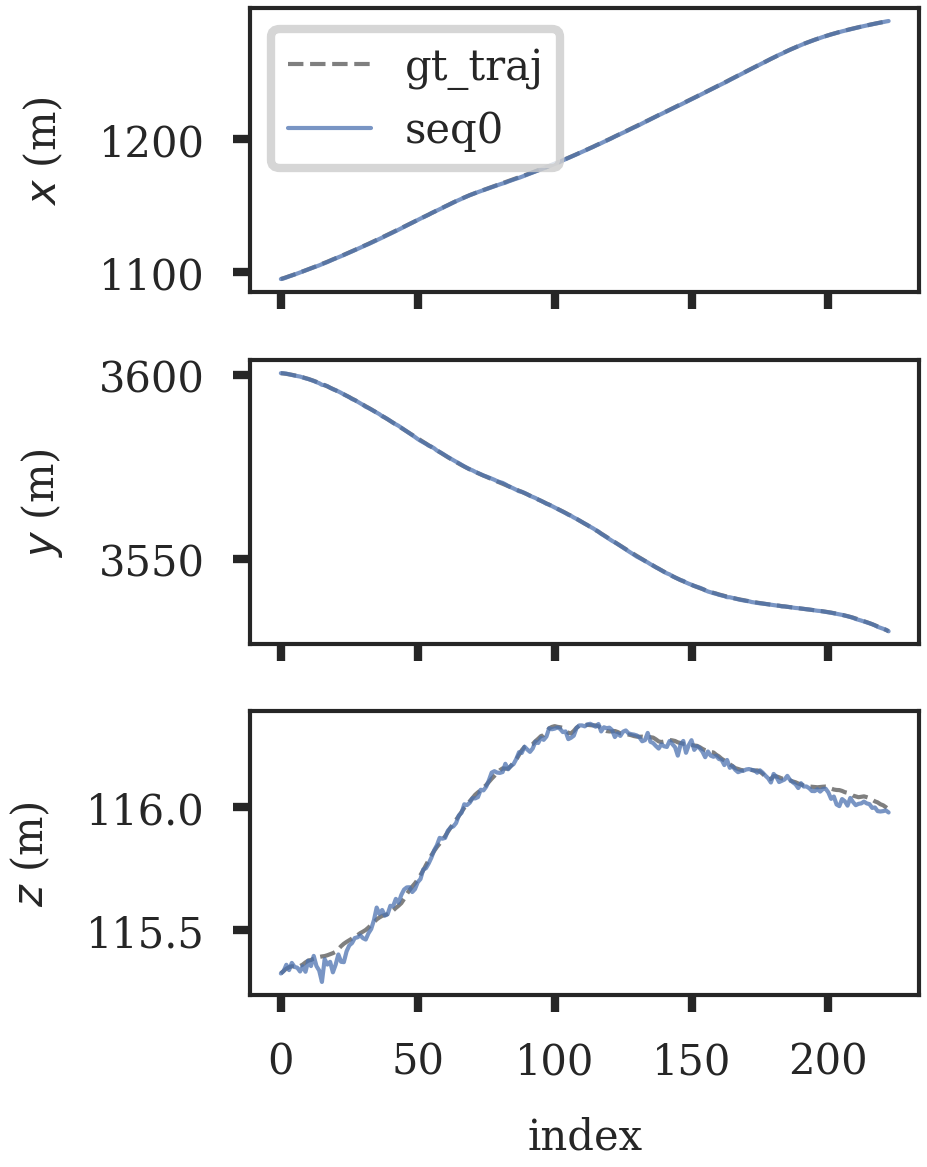}\\
\end{tabular} 
\caption{Comparison of Ground Truth and Predicted Trajectories from Six Views: Roll, Pitch, Yaw, X, Y, Z for Sequence 0} 
\label{fig:seq0_view}
\vspace{\IMGVSPACE}
\end{figure}

\begin{figure}[t]
\vspace{0.2cm}
\centering
\begin{tabular}{cc} 
\includegraphics[width=1.6in]{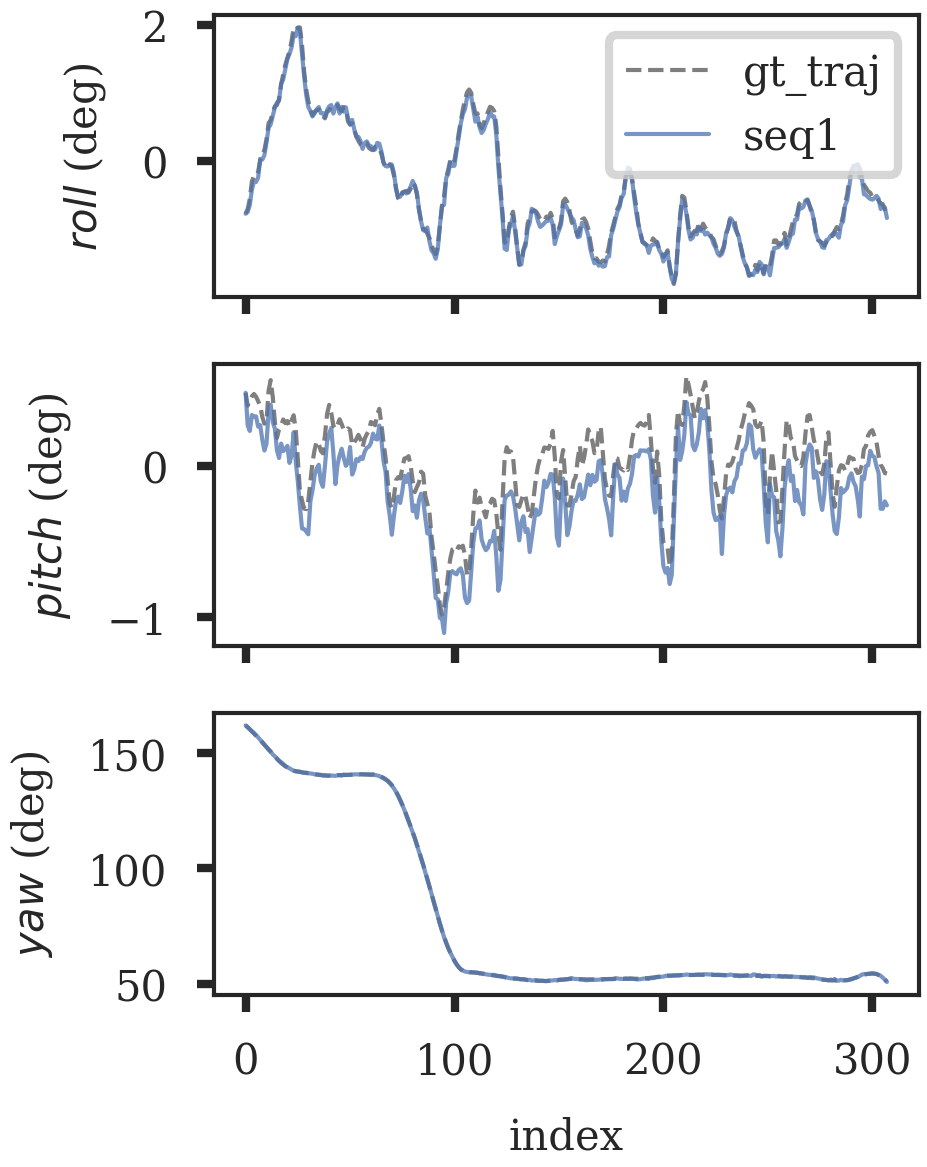}&\includegraphics[width=1.6in]{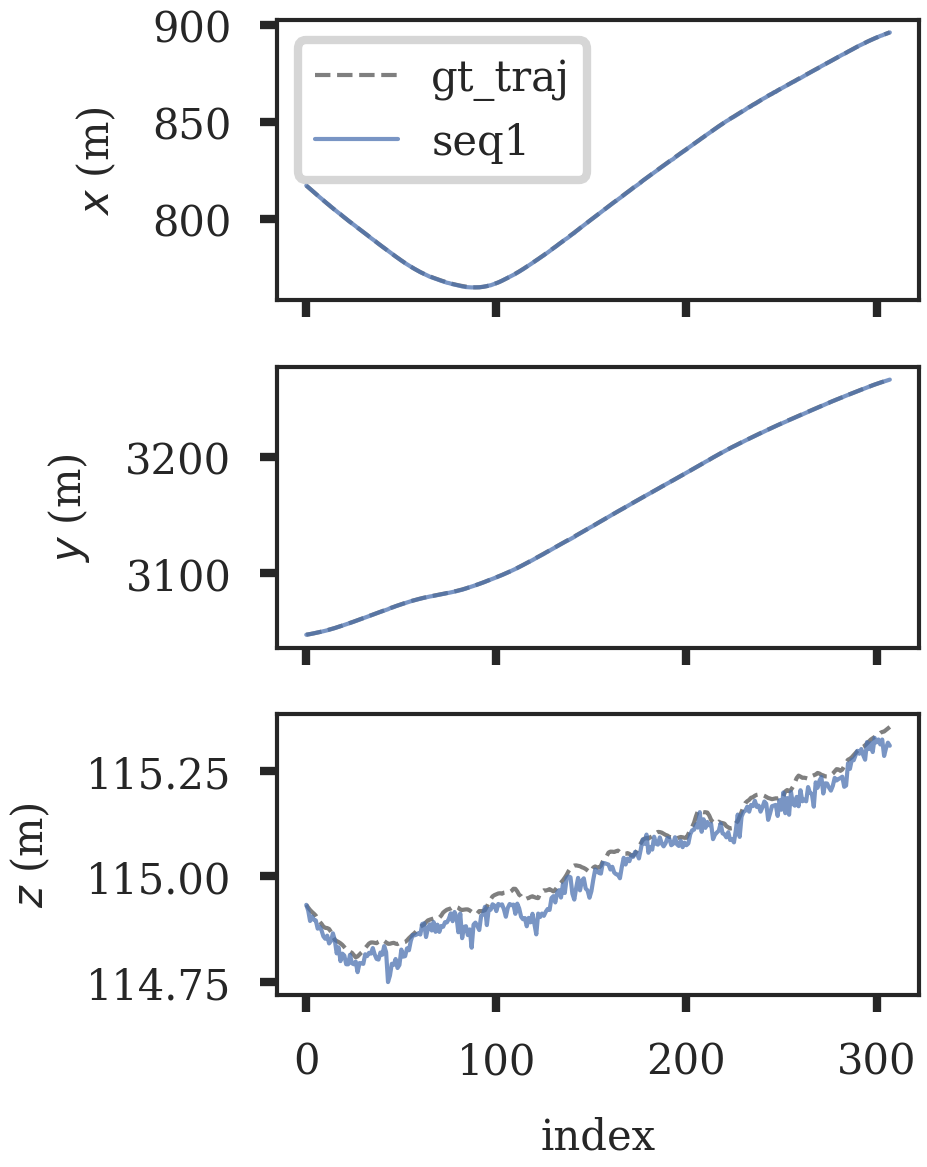}\\
\end{tabular} 
\caption{Comparison of Ground Truth and Predicted Trajectories from Six Views: Roll, Pitch, Yaw, X, Y, Z for Sequence 1} 
\label{fig:seq1_view}
\vspace{\IMGVSPACE}
\end{figure}

Visual analysis of roll-pitch-yaw and XYZ trajectories shows close alignment with ground truth (see Fig. \ref{fig:seq0_view}-\ref{fig:seq1_view}), but pitch and Z-axis estimates have more noise. The noise may stem from inaccuracies in key points extracted from ground features, which are less precisely depicted in 3D Gaussian plots. To enhance accuracy and reduce noise, employing more sophisticated trajectory estimation techniques, like filter-based methods, could offer smoother and more accurate results.
\section{Limitation and Discussion}\label{sec:limit_disc}
\subsection{Balancing Visual Quality with Memory and Geometric Fidelity}\label{sec:dis_vq}
\begin{figure}[t]
\centering
\begin{tabular}{c} 
\includegraphics[width=3.3in]{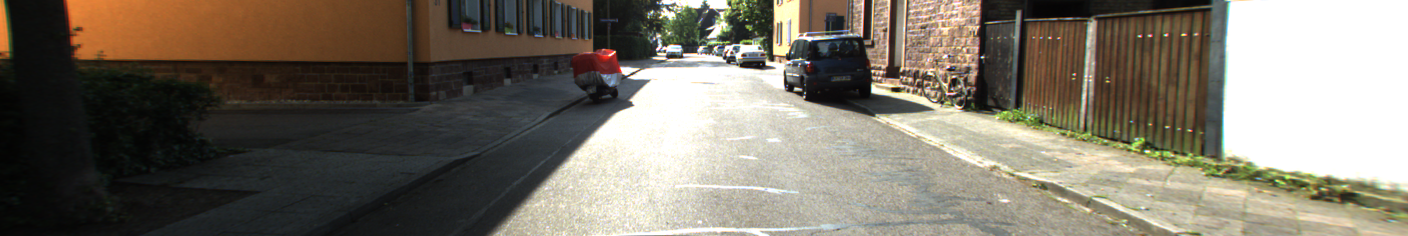}\\
{\scriptsize(a) Ground with Reflection}\\
\includegraphics[width=3.3in]{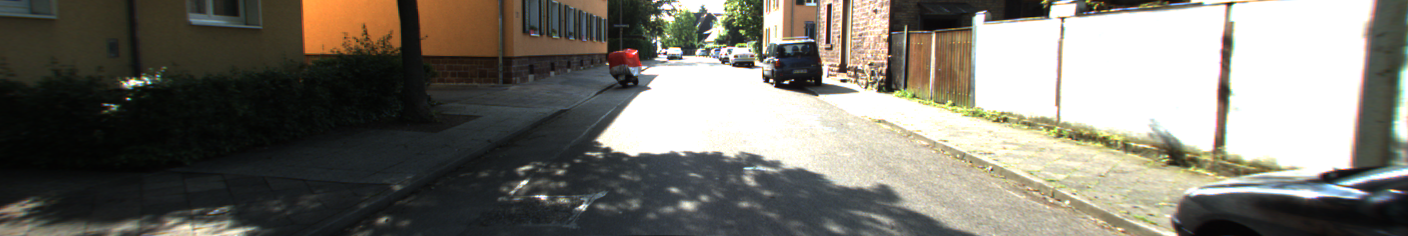}\\
{\scriptsize(b) Ground without Reflection after Moving forward}\\
\includegraphics[width=3.3in]{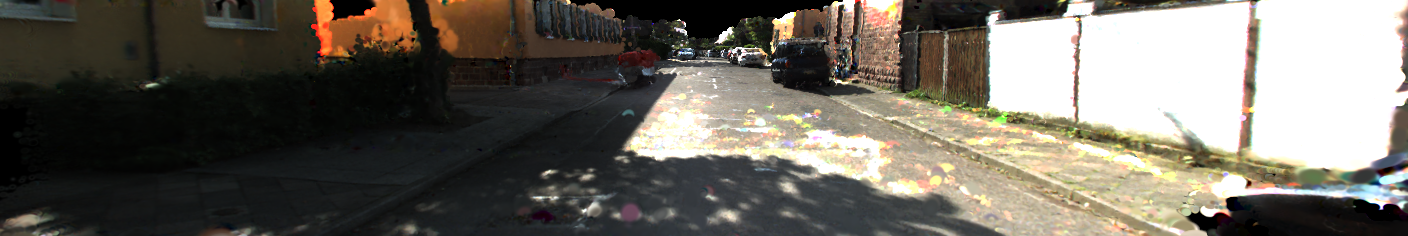}\\
{\scriptsize(c) Rendered Image}\\
\end{tabular} 
\caption{Without encoding lighting information in Gaussian Map, lighting changes cause rendering artifacts} 
\label{fig:light}
\vspace{\IMGVSPACE}
\end{figure}

To minimize the map's footprint, we opted against using Spectral Harmonics (SH) to encode lighting and view-dependent information. While effective in reducing memory usage, this decision has its trade-offs, particularly in outdoor environments where dynamic lighting plays a significant role. For instance, as illustrated in Fig.\ref{fig:light}, changes in lighting direction can result in varying ground colors, leading to artifacts in our rendered images. Despite this challenge, which was particularly noticeable in different Seq 0 due to varying lighting conditions, the localization accuracy between Seq 0 and Seq 1 remained consistent in our experiments. This resilience is primarily attributed to the robustness of the Normalized Cross Correlation (NCC) metrics, as well as the feature detection and matching capabilities of Superpoint\cite{detoneSuperPointSelfSupervisedInterest2018} and LightGlue\cite{lindenbergerLightGlueLocalFeature2023}.
This observation prompts a reevaluation of the necessity to encode lighting information within the map for relocalization tasks. Our findings suggest incorporating dynamic lighting and shadows may not be essential for achieving accurate localization. Moreover, an ideal map might benefit from eliminating lighting and shadow effects to focus more on the environment's geometric and structural aspects, further streamlining the localization process without compromising accuracy. This finding suggests a potential direction for future research, exploring the balance between visual fidelity, memory efficiency, and geometric accuracy in the context of map representation and localization.
\subsection{Towards a Fully Differentiable Localization Pipeline}\label{sec:dis_diff}
The 3D Gaussian Splatting representation's differentiability is an interesting feature, which might offer the possibility of creating a fully differentiable pipeline to perform localization on a 3D Gaussian Splatting submap. This capability might enable us to bypass the traditional detection-description-matching approach, removing the need to train separate models for feature detection and extraction. Additionally, a fully differentiable pipeline can facilitate integration with other differentiable methods for navigation and planning systems. We have initially evaluated several metrics to perform direct localization on a 3D Gaussian Splatting map. These metrics include Gradient Correlation (GC), Normalized Cross Correlation (NCC) \cite{hiasa2018cross}, and Mutual Information (MI) \cite{jiangSemCalSemanticLiDARCamera2021}. However, our preliminary experiment indicates that these metrics are particularly sensitive to initial pose estimates and prone to settling into local minima during gradient descent optimization. These challenges suggest the need to explore alternative optimization techniques or strategies to address these issues.
\section{conclusion}
This paper has explored the integration of LiDAR and camera data through the novel application of 3D Gaussian Splatting, addressing the crucial need for advanced map representation methodologies in the rapidly evolving domains of autonomous driving and robotic navigation. By leveraging the strengths of both LiDAR's depth sensing and the detailed imagery provided by cameras, we have demonstrated a robust approach to creating detailed and geometrically accurate environmental representations, crucial for the safe and efficient navigation of autonomous systems. Our methodology, which begins with LiDAR data to initiate the training of the 3D Gaussian Splatting representation, facilitates the construction of comprehensive maps while addressing common challenges such as high memory usage and inaccuracies in underlying geometry.

The implementation of our technique in visual relocalization task showcases its capacity to enhance the precision of feature identification and positioning, contributing significantly to the field by enabling more sophisticated perception systems for autonomous vehicles. Through our evaluation with the KITTI360 dataset, we have validated the effectiveness, adaptability, and precision of our approach, underscoring its potential to advance environmental perception and system reliability. Ultimately, our work contributes to the broader conversation on sensor data integration and map representation, offering a pathway toward more efficient, accurate, and reliable localization and navigation in complex urban landscapes.




\bibliographystyle{IEEEtran}
\bibliography{IEEEabrv,Reference}

\begin{thebibliography}{10}
\providecommand{\url}[1]{#1}
\csname url@rmstyle\endcsname
\providecommand{\newblock}{\relax}
\providecommand{\bibinfo}[2]{#2}
\providecommand\BIBentrySTDinterwordspacing{\spaceskip=0pt\relax}
\providecommand\BIBentryALTinterwordstretchfactor{4}
\providecommand\BIBentryALTinterwordspacing{\spaceskip=\fontdimen2\font plus
\BIBentryALTinterwordstretchfactor\fontdimen3\font minus \fontdimen4\font\relax}
\providecommand\BIBforeignlanguage[2]{{%
\expandafter\ifx\csname l@#1\endcsname\relax
\typeout{** WARNING: IEEEtran.bst: No hyphenation pattern has been}%
\typeout{** loaded for the language `#1'. Using the pattern for}%
\typeout{** the default language instead.}%
\else
\language=\csname l@#1\endcsname
\fi
#2}}

\bibitem{kerbl3DGaussianSplatting2023}
B.~Kerbl, G.~Kopanas, T.~Leimkühler, and G.~Drettakis, ``{{3D Gaussian Splatting}} for {{Real-Time Radiance Field Rendering}},'' vol.~42, no.~4, p.~1.

\bibitem{liaoKITTI360NovelDataset2023}
Y.~Liao, J.~Xie, and A.~Geiger, ``{{KITTI-360}}: {{A Novel Dataset}} and {{Benchmarks}} for {{Urban Scene Understanding}} in {{2D}} and {{3D}},'' vol.~45, no.~3, pp. 3292--3310.

\bibitem{chenDeepLearningVisual2023}
C.~Chen, B.~Wang, C.~X. Lu, N.~Trigoni, and A.~Markham, ``Deep {{Learning}} for {{Visual Localization}} and {{Mapping}}: {{A Survey}},'' pp. 1--21.

\bibitem{mildenhallNeRFRepresentingScenes2020}
B.~Mildenhall, P.~P. Srinivasan, M.~Tancik, J.~T. Barron, R.~Ramamoorthi, and R.~Ng. {{NeRF}}: {{Representing Scenes}} as {{Neural Radiance Fields}} for {{View Synthesis}}.

\bibitem{sucarIMAPImplicitMapping2021}
E.~Sucar, S.~Liu, J.~Ortiz, and A.~J. Davison, ``{{iMAP}}: {{Implicit Mapping}} and {{Positioning}} in {{Real-Time}},'' pp. 6229--6238.

\bibitem{zhuNICESLAMNeuralImplicit2022}
Z.~Zhu, S.~Peng, V.~Larsson, W.~Xu, H.~Bao, Z.~Cui, M.~R. Oswald, and M.~Pollefeys, ``{{NICE-SLAM}}: {{Neural Implicit Scalable Encoding}} for {{SLAM}}.''

\bibitem{yugayGaussianSLAMPhotorealisticDense2023}
V.~Yugay, Y.~Li, T.~Gevers, and M.~R. Oswald. Gaussian-{{SLAM}}: {{Photo-realistic Dense SLAM}} with {{Gaussian Splatting}}.

\bibitem{matsukiGaussianSplattingSLAM2023}
H.~Matsuki, R.~Murai, P.~H.~J. Kelly, and A.~J. Davison. Gaussian {{Splatting SLAM}}.

\bibitem{liSGSSLAMSemanticGaussian2024}
M.~Li, S.~Liu, H.~Zhou, G.~Zhu, N.~Cheng, and H.~Wang. {{SGS-SLAM}}: {{Semantic Gaussian Splatting For Neural Dense SLAM}}.

\bibitem{keethaSplaTAMSplatTrack2023}
N.~Keetha, J.~Karhade, K.~M. Jatavallabhula, G.~Yang, S.~Scherer, D.~Ramanan, and J.~Luiten. {{SplaTAM}}: {{Splat}}, {{Track}} \& {{Map 3D Gaussians}} for {{Dense RGB-D SLAM}}.

\bibitem{linVastGaussianVast3D2024}
J.~Lin, Z.~Li, X.~Tang, J.~Liu, S.~Liu, J.~Liu, Y.~Lu, X.~Wu, S.~Xu, Y.~Yan, and W.~Yang. {{VastGaussian}}: {{Vast 3D Gaussians}} for {{Large Scene Reconstruction}}.

\bibitem{chenPeriodicVibrationGaussian2023}
Y.~Chen, C.~Gu, J.~Jiang, X.~Zhu, and L.~Zhang. Periodic {{Vibration Gaussian}}: {{Dynamic Urban Scene Reconstruction}} and {{Real-time Rendering}}.

\bibitem{yanStreetGaussiansModeling2024a}
Y.~Yan, H.~Lin, C.~Zhou, W.~Wang, H.~Sun, K.~Zhan, X.~Lang, X.~Zhou, and S.~Peng. Street {{Gaussians}} for {{Modeling Dynamic Urban Scenes}}.

\bibitem{zhouDrivingGaussianCompositeGaussian2024}
X.~Zhou, Z.~Lin, X.~Shan, Y.~Wang, D.~Sun, and M.-H. Yang. {{DrivingGaussian}}: {{Composite Gaussian Splatting}} for {{Surrounding Dynamic Autonomous Driving Scenes}}.

\bibitem{brachmannDSACDifferentiableRANSAC2017}
E.~Brachmann, A.~Krull, S.~Nowozin, J.~Shotton, F.~Michel, S.~Gumhold, and C.~Rother, ``{{DSAC}} - {{Differentiable RANSAC}} for {{Camera Localization}},'' pp. 6684--6692.

\bibitem{laskarCameraRelocalizationComputing2017}
Z.~Laskar, I.~Melekhov, S.~Kalia, and J.~Kannala, ``Camera {{Relocalization}} by {{Computing Pairwise Relative Poses Using Convolutional Neural Network}},'' pp. 929--938.

\bibitem{sarlinBackFeatureLearning2021}
P.-E. Sarlin, A.~Unagar, M.~Larsson, H.~Germain, C.~Toft, V.~Larsson, M.~Pollefeys, V.~Lepetit, L.~Hammarstrand, F.~Kahl, and T.~Sattler, ``Back to the {{Feature}}: {{Learning Robust Camera Localization From Pixels To Pose}},'' pp. 3247--3257.

\bibitem{saputraVisualSLAMStructure2018}
M.~R.~U. Saputra, A.~Markham, and N.~Trigoni, ``Visual {{SLAM}} and {{Structure}} from {{Motion}} in {{Dynamic Environments}}: {{A Survey}},'' vol.~51, no.~2, pp. 37:1--37:36.

\bibitem{wernerTruncatedSignedDistance2014}
D.~Werner, A.~Al-Hamadi, and P.~Werner, ``Truncated {{Signed Distance Function}}: {{Experiments}} on {{Voxel Size}},'' in \emph{Image {{Analysis}} and {{Recognition}}}, ser. Lecture {{Notes}} in {{Computer Science}}, A.~Campilho and M.~Kamel, Eds.\hskip 1em plus 0.5em minus 0.4em\relax Springer International Publishing, pp. 357--364.

\bibitem{wolcottVisualLocalizationLIDAR2014}
R.~W. Wolcott and R.~M. Eustice, ``Visual localization within {{LIDAR}} maps for automated urban driving,'' in \emph{2014 {{IEEE}}/{{RSJ International Conference}} on {{Intelligent Robots}} and {{Systems}}}, pp. 176--183.

\bibitem{detoneSuperPointSelfSupervisedInterest2018}
D.~DeTone, T.~Malisiewicz, and A.~Rabinovich, ``{{SuperPoint}}: {{Self-Supervised Interest Point Detection}} and {{Description}},'' pp. 224--236.

\bibitem{wangP2NetJointDescription2021}
B.~Wang, C.~Chen, Z.~Cui, J.~Qin, C.~X. Lu, Z.~Yu, P.~Zhao, Z.~Dong, F.~Zhu, N.~Trigoni, and A.~Markham, ``P2-{{Net}}: {{Joint Description}} and {{Detection}} of {{Local Features}} for {{Pixel}} and {{Point Matching}}.''

\bibitem{jiangContrastiveLearningFeatures2022b}
P.~Jiang and S.~Saripalli, ``Contrastive {{Learning}} of {{Features}} between {{Images}} and {{LiDAR}},'' in \emph{2022 {{IEEE}} 18th {{International Conference}} on {{Automation Science}} and {{Engineering}} ({{CASE}})}, pp. 411--417.

\bibitem{maxOpticalModelsDirect1995a}
N.~Max, ``Optical models for direct volume rendering,'' vol.~1, no.~2, pp. 99--108.

\bibitem{zwickerEWAVolumeSplatting2001}
M.~Zwicker, J.~Pfister, H.Baar, and M.~Gross, ``{{EWA}} volume splatting,'' in \emph{Proceedings {{Visualization}}, 2001. {{VIS}} '01.}, pp. 29--538.

\bibitem{hiasa2018cross}
Y.~Hiasa, Y.~Otake, M.~Takao, T.~Matsuoka, K.~Takashima, A.~Carass, J.~L. Prince, N.~Sugano, and Y.~Sato, ``Cross-modality image synthesis from unpaired data using cyclegan,'' in \emph{International Workshop on Simulation and Synthesis in Medical Imaging}.\hskip 1em plus 0.5em minus 0.4em\relax Springer, 2018, pp. 31--41.

\bibitem{lindenbergerLightGlueLocalFeature2023}
P.~Lindenberger, P.-E. Sarlin, and M.~Pollefeys, ``{{LightGlue}}: {{Local Feature Matching}} at {{Light Speed}},'' pp. 17\,627--17\,638.

\bibitem{thrunProbabilisticRoboticsIntelligent2005}
S.~Thrun, W.~Burgard, and D.~Fox, \emph{Probabilistic {{Robotics}} ({{Intelligent Robotics}} and {{Autonomous Agents}})}.\hskip 1em plus 0.5em minus 0.4em\relax The MIT Press.

\bibitem{luitenDynamic3DGaussians2023}
J.~Luiten, G.~Kopanas, B.~Leibe, and D.~Ramanan. Dynamic {{3D Gaussians}}: {{Tracking}} by {{Persistent Dynamic View Synthesis}}.

\bibitem{jiangSemCalSemanticLiDARCamera2021}
P.~Jiang, P.~Osteen, and S.~Saripalli, ``{{SemCal}}: {{Semantic LiDAR-Camera Calibration}} using {{Neural Mutual Information Estimator}},'' in \emph{2021 {{IEEE International Conference}} on {{Multisensor Fusion}} and {{Integration}} for {{Intelligent Systems}} ({{MFI}})}.\hskip 1em plus 0.5em minus 0.4em\relax IEEE, pp. 1--7.

\end{thebibliography}

\end{document}